\documentclass{bmvc2k_arxiv}

\title{Iteratively Trained Interactive Segmentation}

\addauthor{Sabarinath Mahadevan}{mahadevan@vision.rwth-aachen.de}{1}
\addauthor{Paul Voigtlaender}{voigtlaender@vision.rwth-aachen.de}{1}
\addauthor{Bastian Leibe}{leibe@vision.rwth-aachen.de}{1}

\addinstitution{
 Computer Vision Group\\
 Visual Computing Institute\\
 RWTH Aachen University\\
 Germany
}

\runninghead{Mahadevan \etal}{Iteratively Trained Interactive Segmentation}

\def\eg{\emph{e.g}\bmvaOneDot}

\def\etal{\emph{et al}\bmvaOneDot}

\def\ie{\emph{i.e}\bmvaOneDot}
\def\cf{\emph{cf}\bmvaOneDot}
\newcommand{\PAR}[1]{\vskip1pt \noindent {\bf #1~}}
\newcommand{\PARbegin}[1]{\noindent {\bf #1~}}
\usepackage{enumitem}

\usepackage{tabularx}
\usepackage[subfigure]{tocloft}
\usepackage{tikz}
\def\checkmark{\tikz\fill[scale=0.2](0,.35) -- (.25,0) -- (1,.7) -- (.25,.15) -- cycle;}

\begin{document}

\maketitle

\begin{abstract}

Deep learning requires large amounts of training data to be effective. For the task of object segmentation, manually labeling data is very expensive, and hence interactive methods are needed. Following recent approaches, we develop an interactive object segmentation system which uses user input in the form of clicks as the input to a convolutional network. While previous methods use heuristic click sampling strategies to emulate user clicks during training, we propose a new iterative training strategy. During training, we iteratively add clicks based on the errors of the currently predicted segmentation. We show that our iterative training strategy together with additional improvements to the network architecture results in improved results over the state-of-the-art.

\end{abstract}
\section{Introduction}
\label{sec:intro}
Recently, deep learning has revolutionized computer vision and has led to greatly improved results across different tasks. However, to achieve optimal performance, deep learning requires large amounts of annotated training data. For some tasks like image classification, manual labels can be obtained with low effort and hence huge amounts of data are available (\eg, ImageNet \cite{Deng09CVPR}). For the task of image segmentation, however, the effort+

Interactive segmentation methods can be based on different kinds of user inputs, such as scribbles or clicks which correct mistakes in the segmentation. In this work, we focus on interactive segmentation of objects using 
clicks as user inputs \cite{Xu16CVPR}. Positive and negative clicks are used by the annotator to add pixels to or to remove pixels from the object of interest, respectively.

Following Xu \etal \cite{Xu16CVPR}, we train a convolutional network which takes an image and some user clicks as input and produces a segmentation mask (see Fig.~\ref{fig:arch} for an overview of the proposed method). Since obtaining actual user clicks for training the network would require significant effort, recent methods \cite{Xu16CVPR, Liew17ICCV} use emulated click patterns. Xu \etal \cite{Xu16CVPR} use a combination of three different heuristic click sampling strategies to sample a set of clicks for each input image during training. At test time, they add clicks one by one and sample the clicks based on the errors of the currently predicted mask to imitate a user who always corrects the largest current error. The strategies applied during training and testing are very different and the sampling strategy for training is  independent of the errors made by the network. We propose to solve this mismatch between training and testing by applying a single sampling strategy during training and testing and demonstrate significantly improved results. We further show that the improvements do not merely result from ``overfitting'' to the evaluation criterion by demonstrating that the results of our method are robust against variations in the click sampling strategy applied at test time.

Additionally, we compare different design choices for representing click and mask inputs to the network. Adopting the state-of-the-art DeepLabV3+ architecture \cite{Chen18arXiv} for our network, we demonstrate that applying the iterative training procedure yields significantly improved results which surpass the state-of-the-art both for interactively creating segmentations from scratch and for correcting segmentations which are automatically obtained by a video object segmentation method.

Our contributions are the following: We introduce Iteratively Trained Interactive Segmentation (ITIS), a framework for interactive click-based image segmentation and make code and models publicly available. As part of ITIS, we propose a novel iterative training strategy. Furthermore we systematically compare different design choices for representing click and mask inputs. We show that ITIS significantly improves the state of the art in interactive image segmentation.
\section{Related Work}
Segmenting objects interactively using clicks, scribbles, or bounding boxes has always been an interesting problem for computer vision research, as it can solve some of the problems in segmentation quality faced by fully-automatic methods. 

Before the success of deep learning, graphical models were popular for interactive segmentation tasks.
Boykov \etal \cite{Boykov01ICCV} use a graph cut based method for segmenting objects in images. In their approach, a user first marks the foreground and background regions in an image which is then used to find a globally optimal segmentation.
Rother \etal proposed an extension of the graph cut method which they call GrabCut \cite{Rother04SIGGRAPH}. 
Here, the user draws a loose rectangle around the object to segment and the GrabCut method extracts the object automatically by an iterative optimisation algorithm. 
Yu \etal \cite{Yu17ICIP} further optimise the results for the problem of interactive segmentation by developing an algorithm called LooseCut.

As with most of the other computer vision algorithms, deep learning based interactive segmentation approaches \cite{Castrejon17CVPR, Xu16CVPR, Liew17ICCV} have recently become popular. Those algorithms learn a strong representation of objectness, \ie which pixels belong to an object and which ones do not. 
Hence, they can reduce the number of user interactions required for generating high  quality annotations.

Lin \etal \cite{Lin16CVPR} 
use scribbles as a form of supervision for a fully convolutional network to segment images. The algorithm is based on a graphical model which is jointly trained with the network.
Xu \etal propose a deep learning based interactive segmentation approach to generate instance segmentations, called iFCN \cite{Xu16CVPR}. iFCN takes user interactions in the form of positive and negative clicks, where a positive click is made on the object that should be segmented (foreground) and a negative click is made on the background. These clicks are transformed into Euclidean distance maps, which are then concatenated with the input channels. 
The concatenated input is fed into a Fully Convolutional Network (FCN) to generate the respective output. Our method is inspired by iFCN but extends it with a recent network architecture and a novel training procedure significantly increasing the  performance.

A more recent work by Liew \etal, called RIS-Net \cite{Liew17ICCV}, uses regional information surrounding the user inputs along with global contextual information to improve the click refinement process. RIS-Net also introduces a click discount factor while training to ensure that a minimal amount of user click information is used and also apply graph cut optimisation to produce the final result.
We show that our method achieves better results without the need for graph cuts and the relatively complicated combination of local and global context. However, these components are complementary and could be combined with our method for further improvements.

In contrast to these methods, which allow clicks at arbitrary positions, DEXTR \cite{Maninis17CVPR} uses extreme points on the objects to generate the corresponding segmentation masks. 
These points are encoded as Gaussians and are concatenated as an extra channel to the image which then serves as an input to a Convolutional Neural Network (CNN).
While this method produces very good results, it has the restriction that exactly four clicks are used for generating the segmentations. It is difficult to refine the generated annotation with additional clicks when this method fails to produce high quality segmentations.

Castrejon \etal \cite{Castrejon17CVPR} propose a Polygon-RNN which predicts a polygon outlining an object to be segmented. The polygons can then interactively be corrected. Their approach shows promising results. However, it requires the bounding box of the object to be segmented as input and it cannot easily be used to correct an existing pixel mask (\eg, obtained by an automatic video object segmentation system) which is not based on polygons. Another disadvantage is that it cannot easily deal with objects with multiple connected components, \eg, a partially occluded car.

\section{Proposed Method}
\begin{figure}[t]
	\begin{center}
		\includegraphics[width=1.0\textwidth]{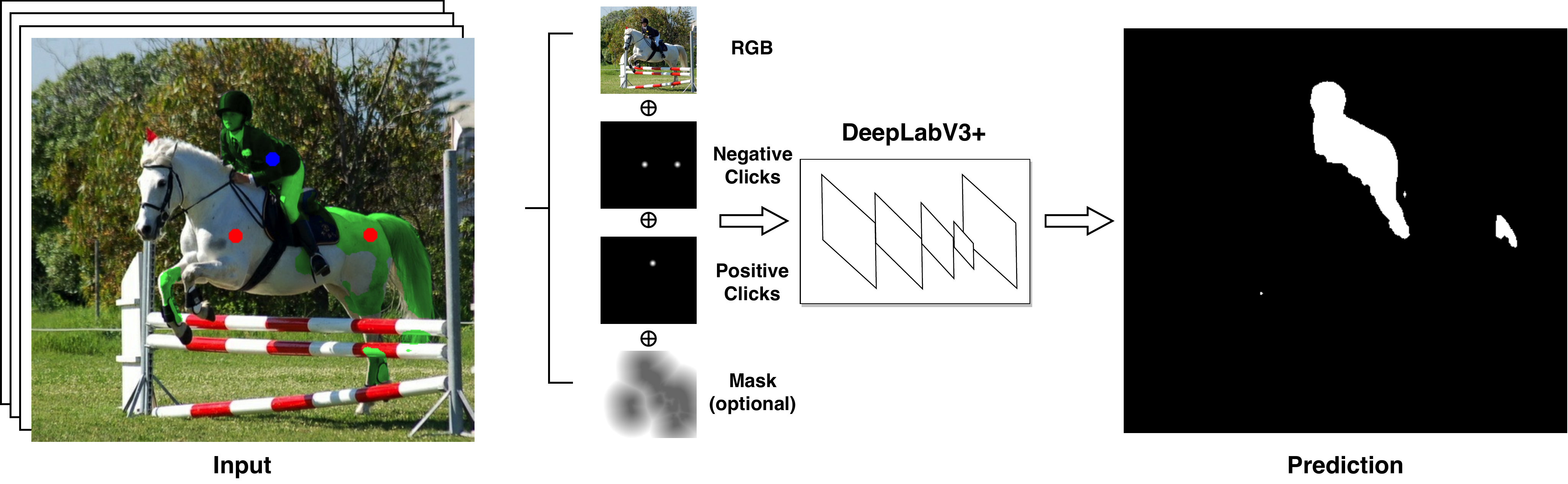}
		\caption{Overview of our method. The input to our network consists of an RGB image concatenated with two click channels representing negative and positive clicks, and also an optional mask channel encoded as distance transform.}
		\label{fig:arch}
	\end{center}
\end{figure}

We propose a deep learning based interactive object segmentation algorithm that uses user inputs in the form of clicks, similar to iFCN \cite{Xu16CVPR}. We propose a novel iterative training procedure which significantly improves the results. Additionally, in contrast to iFCN we encode the user clicks as Gaussians with a small variance and optionally provide an existing estimate of the segmentation mask as additional information to the network.

Figure \ref{fig:arch} shows an overview of our method. We concatenate three additional channels with the input image to form an input which contains six channels in total, where the first two non-color channels represent the positive and the negative clicks, and the (optional) third non-color channel encodes the mask from the previous iteration. 
We use the mask channel only for setups where we have an existing mask to be refined, as we found that when starting from scratch the mask channel does not yield any benefits. When an existing mask is given, we encode it as a Euclidean distance transform, which we found to perform slightly better than using the mask directly as input.
As the name suggests, positive clicks are  made  on  the  foreground,  which in this case is the object to segment, and negative clicks are made on the background. 


\subsection{Network Architecture}
We adopt the recent DeepLabV3+ \cite{Chen18arXiv} architecture for semantic segmentation, which combines the advantages of spatial pyramid pooling and an encoder-decoder architecture. 
The backbone DeepLabV3 \cite{Chen17arXiv} structure acts as an encoder module to which an additional decoder module is added to recover fine structures. 
Additionally, DeepLabV3+ adopts depth-wise separable convolutions 
which results in a faster network with fewer parameters. DeeplabV3+ \cite{Chen18arXiv} produces the state of the art performance on the PASCAL VOC 2012 \cite{Everingham10IJCV} dataset. 

All network weights except for the output layer are initialised with those provided by Chen \etal \cite{Chen18arXiv}, which were obtained by pretraining on the ImageNet \cite{Deng09CVPR}, COCO \cite{Lin14ECCV}, and PASCAL VOC 2012 \cite{Everingham10IJCV} datasets. The output layer is replaced with a two-class softmax layer, which is used to produce binary segmentations. In contrast to iFCN \cite{Xu16CVPR}, we directly obtain a final segmentation by thresholding the posteriors produced by the network at $0.5$ and we do not use any post-processing with graphical models.

\subsection{Iterative Training Procedure}
\label{sec:iterative_training}

In order to keep the training effort manageable, we resort to random sampling for generating clicks. We propose an iterative training procedure, where clicks are progressively added based on the errors in the predictions of the network during training, which closely aligns the network to the actual usage pattern of an annotator. This training procedure boosts the performance of our interactive segmentation model, as shown later in the experiments. 

For the iterative training, we use two different kinds of sampling techniques, 
one for obtaining an initial set of clicks and another for adding additional correction clicks based on errors in the predicted masks.
Here, the initialisation strategy helps the network to learn different notions such as negative objects or object boundaries, while the second strategy is useful for learning to correct errors in the predicted segmentation mask. Both strategies help the network learn properties which are useful for an interactive segmentation system.


The training starts for each object with click channels which are initialised with clicks sampled randomly based on the initial click sampling strategy as detailed below. The optional mask channel is initialised to an empty mask, if it is used.



When starting a new epoch, one of the click channels (either positive or negative) is updated with a new correction click which is sampled based on the misclassified pixels in the predicted mask from the last epoch, according to the iterative click addition algorithm (see below).
When adding one click per epoch to each object, after some time the network would only see training examples with many existing clicks and a mask which is already at a high quality. This would degrade the performance for inputs with only few clicks or low quality masks. To avoid this behaviour, at the beginning of an epoch for each object the clicks are reset with probably $p_r$ to a new set of clicks sampled using the initial click sampling strategy (described below). When using the optional mask channel, it is then also reset to an empty mask.
The reset of the clicks also introduces some randomness within the training data which reduces over-fitting.

In the following, we describe the initial click sampling and iterative click addition algorithms.



\PAR{Initial Click Sampling.}
For this, we initialise the click channels using multiple sampling strategies that try to reproduce the click patterns of a human annotator during training, as done in iFCN \cite{Xu16CVPR}. Briefly, iFCN samples positive clicks on the object, and negative clicks based on three different strategies which try to cover multiple patterns such as encoding the object boundary or removing false-positive predictions from background objects. For more details on the sampling strategies, we refer the reader to our supplementary material.

\PAR{Iterative Click Addition.}
After the initial set of clicks are sampled using the above strategies, we generate all subsequent clicks for an image with respect to the segmentation mask which was predicted by the network at the previous iteration, as explained below

\begin{itemize}[noitemsep,topsep=0pt,parsep=0pt,partopsep=0pt]
\item First, the mislabelled pixels from the output mask of the previous iteration $m_{i-1}$ are identified by comparing the output mask with the ground truth mask (see Fig.~\ref{fig:corr_click_strategy} a). 
\item These pixels are then grouped together into multiple clusters using connected component labelling (see Fig.~\ref{fig:corr_click_strategy} b).
	
\item The largest of these clusters is selected based on the pixel count. 
	
\item A click is sampled on the largest cluster (see Fig.~\ref{fig:corr_click_strategy} c) such that the sampled pixel location has the maximum Euclidean distance from both the cluster boundary and the other click points within the same cluster. This corresponds to the centre of the cluster if no previous clicks were sampled on it. Here sampling is only used to break ties if multiple pixels have the same distance.
	
\item Finally, the sampled click is considered as positive if the corresponding pixel location in the target image lies on the object, or as negative otherwise. A Gaussian is subsequently added to the corresponding click channel at the sampled location.


\begin{figure}[t]
			\begin{center}
			\subfigure[       ]{\label{fig:a}\includegraphics[width=0.33\textwidth, trim = 0 0 390 0 , clip]{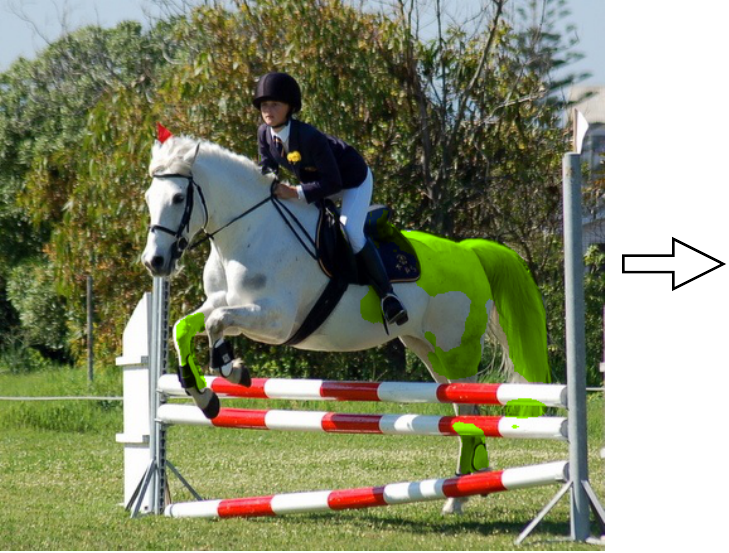}}
\subfigure[       ]{\label{fig:b}\includegraphics[width=0.33\textwidth, trim = 210 0 175 0 , clip]{images/click_sampling}}
\subfigure[       ]{\label{fig:b}\includegraphics[width=0.27\textwidth, trim = 423 0 0 0 , clip]{images/click_sampling}}
				\caption{An example of the proposed click sampling strategy. a) From all mislabeled pixels (shown in green), b) clusters of mislabeled pixels are identified. c) A click is added on the largest mislabelled cluster after each training round.}
				\label{fig:corr_click_strategy}
			\end{center}
			\end{figure}
\end{itemize}

\section{Experiments}
We conduct experiments on four different datasets and compare our approach to other recent methods. 
On the PASCAL, GrabCut, and KITTI datasets, we consider a scenario where objects are segmented using clicks from scratch. 
On the DAVIS dataset, we start with the results obtained by an automatic method for video object segmentation and correct the worst results using our method.
\subsection{Datasets}
\PARbegin{PASCAL VOC.} We use the 1,464 training images from the PASCAL VOC 2012 dataset \cite{Everingham10IJCV} plus the additional instance annotations from the semantic boundaries dataset (SBD) \cite{Hariharan11ICCV} provided by Hariharan \etal for training our network. This provides us with more than 20,000 object instances across 20 categories. For our experiments, we use all 1,449 images of the validation set.

\PAR{GrabCut.} The GrabCut dataset \cite{Rother04SIGGRAPH} consists of 50 images with the corresponding ground truth segmentation masks and is used traditionally by interactive segmentation methods. We evaluate our algorithm on GrabCut to compare our method with other interactive segmentation algorithms.

\PAR{KITTI.} For the experiments on KITTI \cite{Geiger12CVPR}, we use 741 cars annotated at the pixel level provided by \cite{Chen14CVPR}.

\PAR{DAVIS.} DAVIS \cite{Perazzi16CVPR} is a dataset for video object segmentation.  It consists of 50 short videos from which 20 are in the validation set which we use in our experiments. In each video, the pixel masks of all frames for one object are annotated.
\subsection{Experimental Setup}
\label{sec:exp_setup}
For training our network, we use bootstrapped cross-entropy \cite{Wu16arXiv} as the loss function, which takes an average over the loss values at the pixels that represent the worst $k$ predictions. We train on the worst $25\%$ of the pixel predictions and use Adam \cite{Kingma15ICLR} to optimize our network. 
We use a reset probability $p_r$ of $0.3$ (\cf Section \ref{sec:iterative_training}).
The clicks are encoded as Gaussians with a standard deviation of 10 pixels that are centred on each click. We clip the Gaussians to 0 at a distance of 20 pixels from the clicks. Using Gaussians with a small scale localises the clicks well and boosts the system performance, as shown in our experiments. Training is always performed on PASCAL VOC for about 20 epochs. More details are given in the supplementary material.

We use 
the mean intersection over union score (mIoU) calculated between the network prediction and the ground truth, to evaluate the performance of our interactive segmentation algorithm. For a fair comparison with the other interactive segmentation methods, we also report the number of clicks used to reach a particular mIoU score. For this, we run the same setup that is used in other interactive segmentation methods \cite{Liew17ICCV, Xu16CVPR}, where clicks are simulated automatically to correct an existing segmentation. 
The algorithm used to sample the clicks is the same as for iterative click addition during training (\cf \ref{sec:iterative_training}).
Clicks are repeatedly added until 20 clicks are sampled, and the mIoU score is calculated against the number of clicks that are sampled to achieve it. If a particular IoU score cannot be reached for an instance, then the number of clicks is thresholded to 20 \cite{Xu16CVPR}.
\subsection{Comparison to State of the Art}
\label{sec:comp_soa}
\begin{figure}[t]
\centering     
\subfigure[PASCAL VOC]{\label{fig:a}\includegraphics[width=0.49\textwidth, trim = 100 30 100 95 , clip]{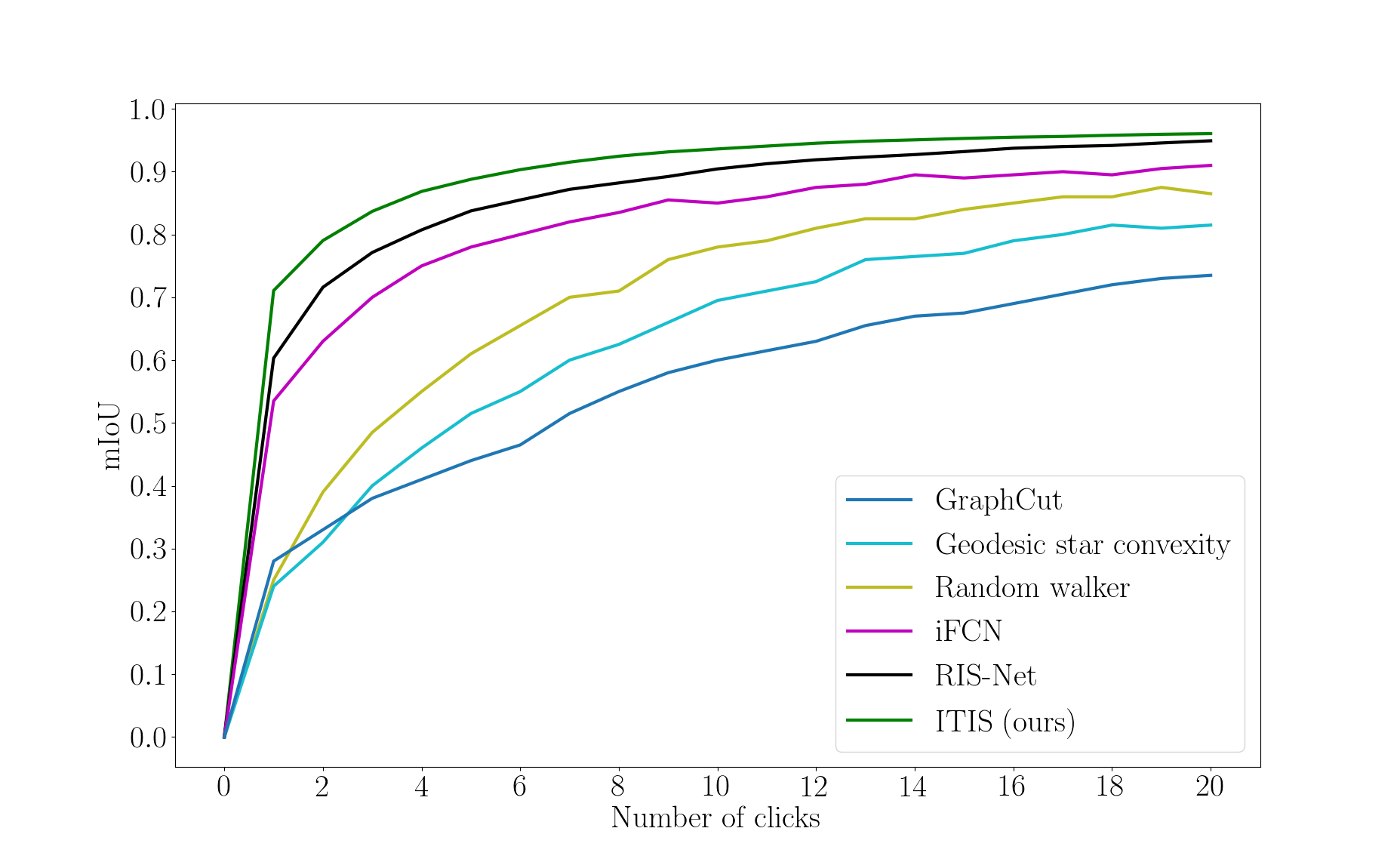}}
\subfigure[GrabCut]{\label{fig:b}\includegraphics[width=0.49\textwidth, trim = 100 30 100 90 , clip]{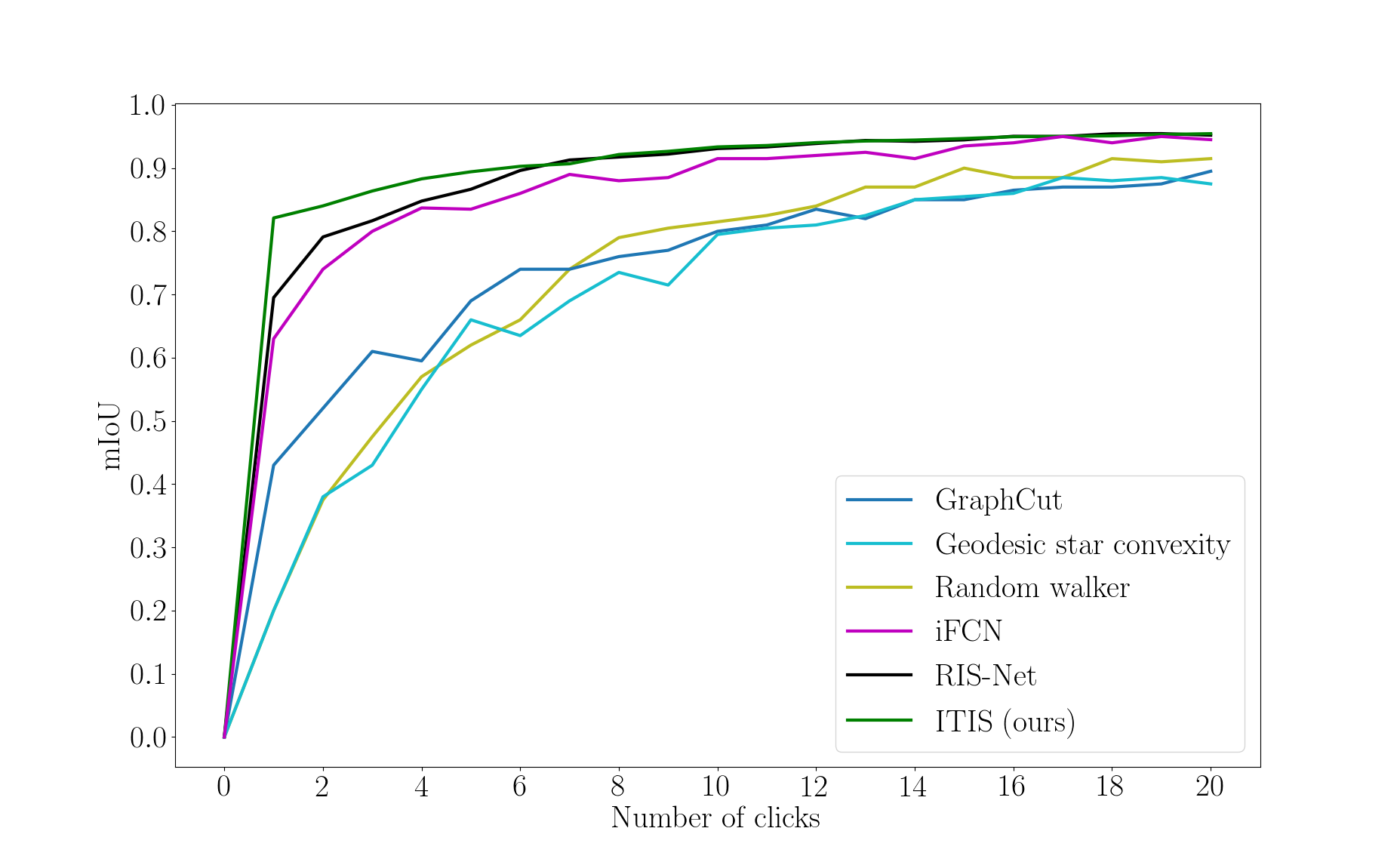}}
\caption{Mean IoU score against the number of clicks used to achieve it on the PASCAL VOC \cite{Everingham10IJCV} and GrabCut \cite{Rother04SIGGRAPH} datasets.}
\label{fig:clicks_vs_iou}
\end{figure}

We consider two different methodologies to evaluate the amount of interaction required to reach a specific mIoU value. 
The first method is to count for each object individually, how many clicks are required to obtain a specific IoU value. If the IoU value cannot be achieved within 20 clicks, the number of clicks for this object is clipped to 20 \cite{Xu16CVPR}. The results for PASCAL and GrabCut using this method are shown in Table \ref{tab:pascal_grabcut_eval} (left). It can be clearly seen from the results that our iteratively trained model requires the least number of clicks on the PASCAL VOC validation set giving a huge advantage over the previous state-of-the-art interactive segmentation methods. An interesting observation here is that our model requires 0.2 clicks less than DEXTR \cite{Maninis17CVPR}, which in fact requires all four extreme points for segmenting an object and hence requires much more human effort compared to our method.
Figure \ref{fig:clicks_vs_iou} a) complements our observations by showing that our model consistently outperforms the other methods on the PASCAL VOC dataset \cite{Everingham10IJCV}.
The curve in Figure \ref{fig:clicks_vs_iou} b) shows however, that our method produces the best result for the initial few clicks and afterwards performs similar to RIS-Net \cite{Liew17ICCV}. To reach the high threshold of $90\%$ on GrabCut \cite{Rother04SIGGRAPH}, our method needs slightly more clicks than RIS-Net \cite{Liew17ICCV}. However we argue that this is mainly an effect of the very high threshold which is for many instances very slightly not reached.

The second way of evaluation is to use the same number of clicks for each instance and to increase the number of clicks until the target IoU value is reached. The results for this evaluation are shown in Table \ref{tab:pascal_grabcut_eval} (right). 
With this evaluation strategy, ITIS performs slightly better than RIS-Net \cite{Liew17ICCV} on GrabCut and again shows the strongest result on the PASCAL VOC \cite{Everingham10IJCV} dataset.

\begin {table}[t]
\footnotesize
\begin{center}
	\resizebox{\columnwidth}{!}{
 \begin{tabular}{||c c c c||} 
 \hline
 Method & PASCAL & GrabCut &\\ 
               &  @ 85\% & @90\%&\\
 \hline\hline
 Graph cut \cite{Boykov01ICCV} & 15.0 & 11.1 &  \\ 
 Geodesic matting \cite{Bai07ICCV}& 14.7 & 12.4 & \\
 Random walker \cite{Grady06TPAMI} & 11.3 & 12.3 & \\
 iFCN\cite{Xu16CVPR} & 6.8 & 6.0 & \\ 
 RIS-Net\cite{Liew17ICCV} & 5.1 & 5.0 &  \\
 DEXTR \cite{Maninis17CVPR}& 4.0 & \textbf{4.0} &  \\
 \hline
 ITIS (ours) & \textbf{3.8} & 5.6 &  \\ [1ex] 
 \hline
 \end{tabular}
\begin{tabular}{||c c c c||} 
	\hline
	Method & PASCAL & GrabCut &\\ 
	&  @ 85\% & @90\%&\\
	\hline\hline
	Graph cut \cite{Boykov01ICCV} & >20 & >20 &\\
	Geodesic matting \cite{Bai07ICCV}& >20 & >20 & \\
	Random walker \cite{Grady06TPAMI} & 16.1 & 15 & \\
	iFCN\cite{Xu16CVPR} & 8.7 & 7.5 & \\ 
	RIS-Net\cite{Liew17ICCV} & 5.7 & 6.0 &  \\
	DEXTR \cite{Maninis17CVPR}& 4.0 & \textbf{4.0} &  \\
	\hline
	ITIS (ours) & \textbf{3.4} & 5.7 &  \\[1ex] 
	\hline
\end{tabular}
}
 \end{center}
 \caption{\label{tab:pascal_grabcut_eval}The average number of clicks required to attain a particular mIoU score on PASCAL VOC 2012 and GrabCut datasets. The table on the left shows the values calculated per object instance, and the one on the right shows the corresponding values over the whole validation set.}
\hfill
\end{table}
\subsection{Generalisation to Other Sampling Strategies}
\begin{figure}[t]
			\begin{center}
				\includegraphics[width=0.4\textwidth, trim = 80 30 100 20 , clip]{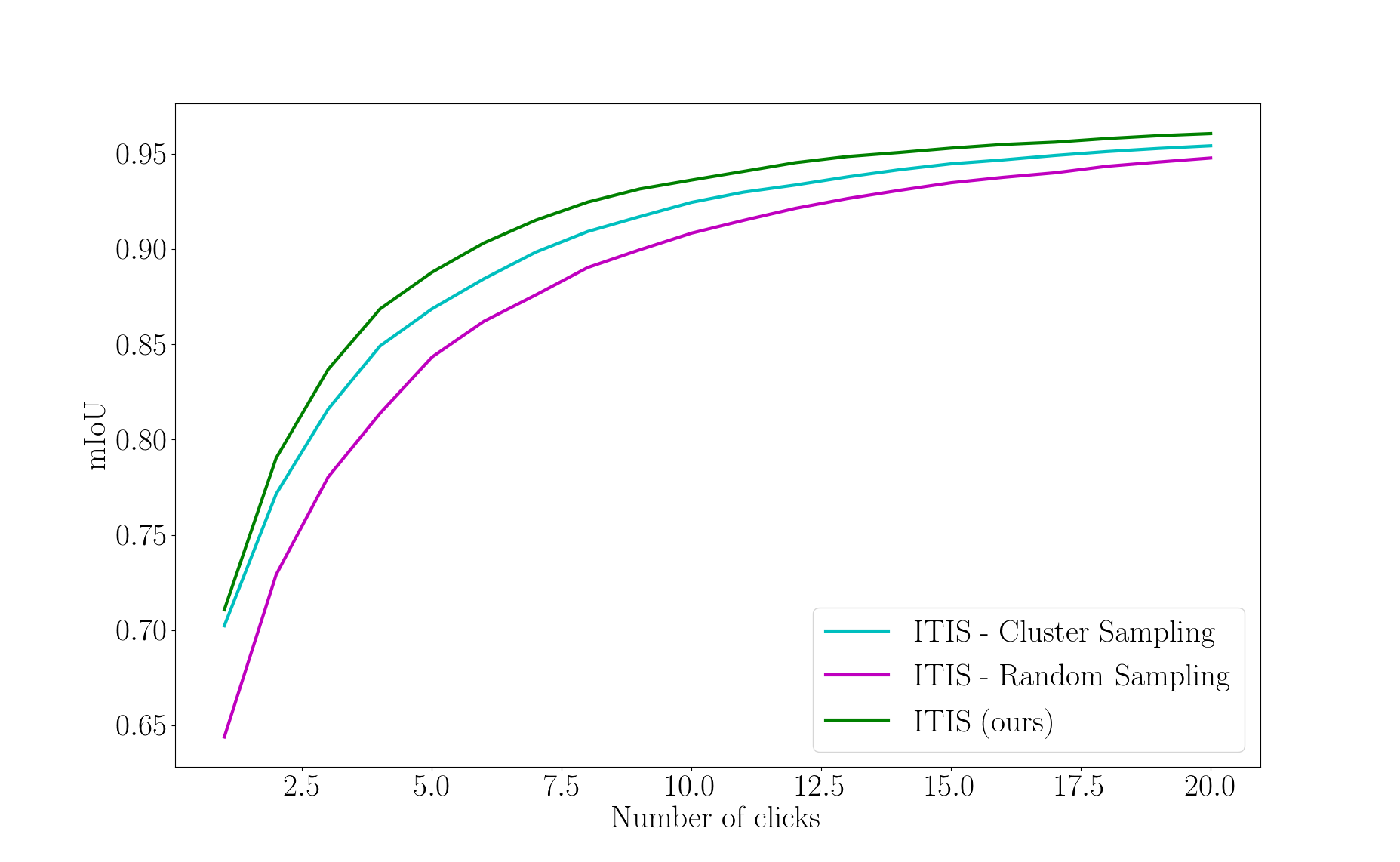}
				\caption{Effect of different click sampling strategies at test time. It can be seen that our method generalizes to alternative sampling methods with only a small loss in performance.}
				\label{fig:generalised_sampling}
			\end{center}
			\end{figure}

To show that our training strategy does not overfit to one particular sampling pattern, we evaluate our method with different click sampling strategies at test time. For this, we use two additional click sampling strategies for correcting the segmentation masks, which we call \textit{cluster sampling}, and \textit{random sampling}. In probability cluster sampling, first the set of mislabelled clusters is identified using connected components labelling, as described in Section \ref{sec:iterative_training}. 
A cluster is then chosen based on a probability proportional to the size of the cluster, and a click is added to the centre of this cluster. For the Random Sampling strategy, we consider the whole misclassified region as a single cluster, and randomly sample a pixel from it. Figure \ref{fig:generalised_sampling} shows the results of our methods with all three sampling strategies. Although smarter sampling strategies, such as cluster sampling, or choosing the largest mislabelled cluster has some advantages for lower number of clicks, this gets neutralised as more clicks are added. The plot shows that our method can achieve similar mIoU scores even with a totally random click sampling strategy further demonstrating that ITIS is robust against user click patterns.

\subsection{Ablation Study}
\begin{figure}
	\begin{minipage}{0.5\textwidth}
			\footnotesize
			\begin{center}
				\resizebox{\columnwidth}{!}{
					\begin{tabular}{||c c | c c | c||}
						\hline
						Distance Transform & Gaussian & Iterative Training & Mask & Clicks\\ [0.5ex] 
						\hline\hline
						\checkmark &  &  &  & 6.5\\ 
						& \checkmark &  &  & 5.4\\
						& \checkmark & \checkmark &  & 3.8\\
						& \checkmark & \checkmark & \checkmark & 3.7\\ [1ex] 
						\hline
					\end{tabular}
				}
			\end{center}
			
	\end{minipage}
    \begin{minipage}{0.5\textwidth}
    		\begin{center}
    			\includegraphics[width=0.7\textwidth, trim = 80 30 100 20 , clip]{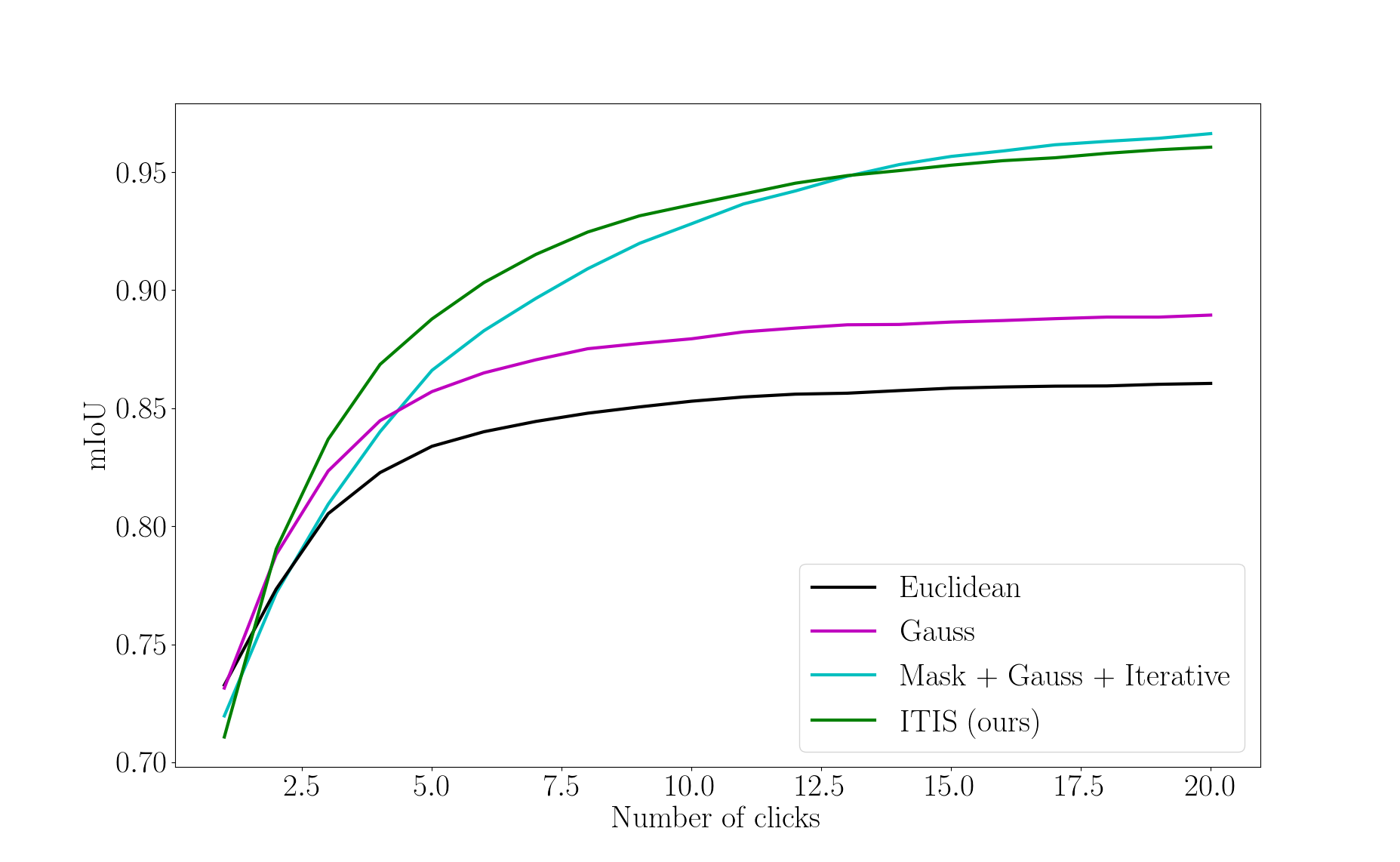}
    		\end{center}
    	\label{fig:ablation_plot}
    \end{minipage}
\caption{\label{tab:ablation} Ablation study on PASCAL VOC. It can be seen, both from the table on the left and the plot on the right, that the proposed iterative training procedure significantly improves the results.}
\end{figure}

We compare different variants of the proposed method in Figure \ref{tab:ablation}. In particular, we investigate the effect of the representation of the clicks on the PASCAL VOC \cite{Everingham10IJCV} dataset. We use the same evaluation strategy as in Section \ref{sec:exp_setup} to compare the different models. iFCN \cite{Xu16CVPR} uses a distance transform to encode clicks, while DEXTR \cite{Maninis17CVPR} and Benard \etal \cite{Benard18arXiv} found that encoding clicks by Gaussians yields better results. Our results also confirm this finding: When we replace the Gaussians by a distance transform, the number of clicks that is required increases from 5.4 to 6.5. The table on the left of Figure \ref{tab:ablation} also shows that the iterative training strategy greatly reduces the number of clicks needed to reach 85\% mIoU on PASCAL VOC from $5.4$ to $3.8$ clicks. When the optional mask channel is added, which in our case is used to evaluate video object segmentation, the model performs on similar lines in terms of the click evaluation. However, this
reduces the performance for the initial 10 clicks as seen in Figure \ref{tab:ablation} (right). It is also worthwhile to note that the iterative training scheme boosts the maximum mIoU achieved by the model at 20 clicks.  

\subsection{Correcting Masks for Video Object Segmentation}
Many recent works \cite{Caelles17CVPR, Perazzi17CVPR, Voigtlaender17BMVC, Khoreva17CVPRW} focus on segmenting objects in videos since such object annotations are expensive. These fully-automatic methods produce results which are of good quality but still contain some errors. In this scenario, we are given existing segmentation masks with errors, which can then be corrected by our method using additional clicks. In order to account for the existing mask, we use the optional mask channel as input to the network in this setting.
Following \cite{Benard18arXiv}, we refine the results obtained by OSVOS \cite{Caelles17CVPR} and report the segmentation quality at 1, 4 and 10 clicks in Table \ref{tab:vos}. The table shows that our extended network, referred to as ITIS - VOS, produces better results compared to the other methods, especially at clicks 1 and 4.

\begin {table}[t]
\footnotesize
\begin{center}
 \begin{tabular}{||c c c c c||} 
 \hline
 Method & OSVOS & 1 click & 4 clicks & 10 clicks \\ [0.5ex] 
 \hline\hline
 GrabCut\cite{Liew17ICCV} & 50.4 & 46.6  & 53.5 & 68.8\\
 iFCN\cite{Xu16CVPR} & 50.4 & 55.7 & 71.3 & 79.9\\ 
 IVOS \cite{Benard18arXiv}& 50.4 & 63.8  & 75.7 & 82.2\\
 \hline
 ITIS - VOS (ours) & 50.4 & \textbf{67.0} & \textbf{77.1} & \textbf{82.8}\\ [1ex] 
 \hline
 \end{tabular}
 \end{center}
 \caption{\label{tab:vos}Refinement of the worst predictions from OSVOS \cite{Caelles17CVPR} (performance measured in \% mIoU). Our method with an additional mask channel refines the predictions significantly with a few number of clicks.}
\hfill
\end{table}
\subsection{Annotating KITTI Instances}
\begin{figure}[t]
\begin{center}
\includegraphics[width=0.45\textwidth]{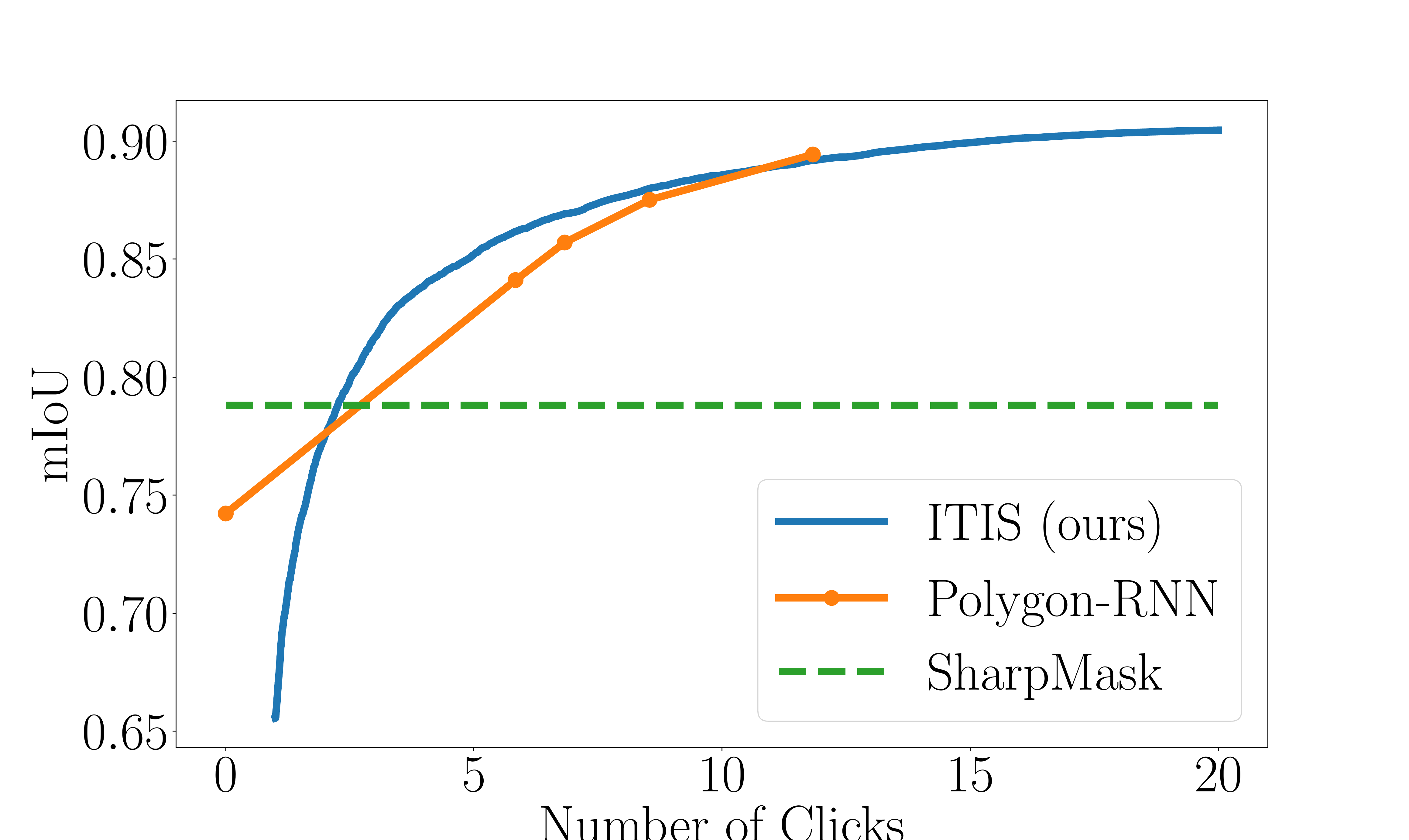}
\caption{Interactive segmentation performance for segmenting 741 cars on KITTI. For a large range of number of clicks our method performs better than Polygon-RNN although Polygon-RNN uses the ground truth bounding box and requires more manual effort per click.}
\label{fig:KITTI}
\end{center}
\end{figure}

In order to compare to Polygon-RNN and to show that our method generalizes to other datasets, we segment 741 cars on the KITTI dataset. The results are shown in Fig. \ref{fig:KITTI}, where we also added the result of the fully-automatic SharpMask \cite{Pinheiro16ECCV} method for comparison. For the results from Polygon-RNN, clicks are added until all vertices are closer than a specific threshold to their ground truth positions. To create a comparable setup, we define an IoU threshold which should be reached per instance and add up to 20 clicks to each instance until the IoU value is reached. 
We then vary the target IoU to generate a curve. Note that the shown mIoU in the curve is not the used threshold, but the actual obtained value.
Polygon-RNN needs the ground truth bounding box in order to crop the instance which allows it to already produce reasonable results at 0 clicks. In contrast, we work on the whole image without needing the bounding box, which in turn means that ITIS takes a couple of clicks to catch up with Polygon-RNN and from there performs better than Polygon-RNN, converging to a similar value for many clicks.
Additionally, correcting a polygon by a click requires significant effort since the click needs to be exactly at the outline of the object while for our method the user just needs to click somewhere in a region which contains errors. Moreover, Polygon-RNN was trained on the Cityscapes \cite{Cordts16CVPR} dataset, with an automotive setup closer to KITTI, while we focus on a generic model trained on Pascal VOC.
\section{Conclusion}
We introduced ITIS, a framework for interactive click-based segmentation with a novel iterative training procedure. We have demonstrated results better than the current state-of-the-art on a variety of tasks. We will make our code including an annotation tool publicly available and hope that it will be used for annotating large datasets. 

\vspace{0.5cm}
\PARbegin{Acknowledgements.}
This project was funded, in parts, by ERC Consolidator 
Grant DeeVise (ERC-2017-COG-773161) and EU project CROWDBOT 
(H2020-ICT-2017-779942). We would like to thank Istv\'{a}n S\'{a}r\'{a}ndi for helpful discussions.

\clearpage
\bibliography{abbrev_short,paper}

\clearpage
\appendix
\textcolor{bmv@sectioncolor}{\vspace{-1.2cm} \part*{Supplementary Material}}
\normalsize

\section{Initial Click Sampling}

To initialise the click channels, we use the click sampling strategies proposed by \cite{Xu16CVPR}. The sampling algorithm works as follows.
	
	\PAR{Positive clicks.} First, the number of positive clicks $n_{pos}$ is sampled from $[1, N_{pos}]$. Then, $n_{pos}$ clicks are randomly sampled from the object pixels, which can be obtained from the ground truth mask. Each of these clicks are sampled such that any two clicks are $d_s$ pixels away from each other and $d_m$ pixels away from the object boundary.
	\PAR{Negative clicks.} For sampling negative clicks, we use multiple strategies to encode the user click patterns. Let us define a strategy set $S=\{s_1, s_2, s_3\}$. First, a strategy is randomly sampled from set $S$ and then the sampled strategy is used to generate $n_{neg}$ clicks on the input image. Here, $n_{neg}$ is a number sampled from $[0, N_i]$ where $i \in [1,2,3]$ and $N_i$ represents the maximum number of clicks for each strategy. The strategies used here are explained in detail below.
	\begin{itemize}[noitemsep,topsep=0pt,parsep=0pt,partopsep=0pt]
	\item \textbf{$s_1$:} In the first strategy, $n_1$ clicks are sampled randomly from the background pixels such that they are within a distance of $d_o$ pixels from the object boundary. The clicks are filtered in the same way as the positive clicks.
	\item \textbf{$s_2$:} The second strategy is to sample $n_2$ clicks on each of the negative objects. Here again, the clicks are filtered to honour the same constraints as in the first strategy.
	\item \textbf{$s_3$:} Here, $N_3$ clicks are sampled to cover the object boundaries. This helps to train the interactive network faster.
	\end{itemize}
	
\section{Implementation Details}

We train with a fixed crop size of $350\times350$ pixels. Input images whose smaller side is less than $350$ pixels are bilinearly upscaled such that the smaller side is $350$ pixels long. Otherwise, the image is kept at the original resolution. Afterwards, we take a random crop which is constrained to contain at least a part of the object to be segmented. The only form of data augmentations we use are gamma augmentations \cite{Pohlen17CVPR}.
We start with a learning rate of $10^{-5}$ and reduce it to $10^{-6}$ at  epoch $10$ and to $3 \cdot 10^{-7}$ at epoch 15. At test time, we use the input image in the original resolution without resizing or cropping.

For the initial click sampling, we set the hyperparameters to $N_{pos}=5$, $d_m=5$, $d_s=40$, $d_o=40$, $N_1=10$, $N_2=5$, $N_3=10$.

\section{Qualitative Results}

\begin{figure}[h!]
\centering     
\subfigure[\textbf{Single click results}. In many cases, ITIS produces good quality segmentations even with a single click.]{\includegraphics[width=\textwidth, trim = 0 330 0 0,clip]{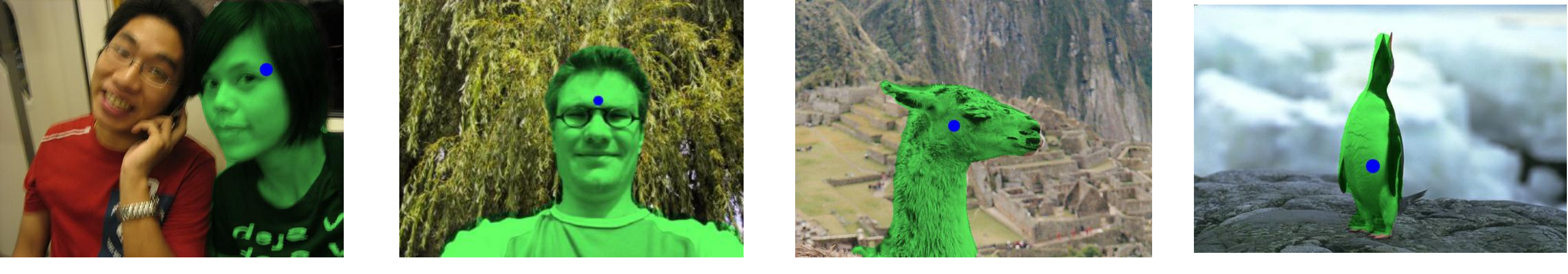}}
\subfigure[\textbf{Multi-click results}. With a few clicks, undesired objects can be removed.]{\label{fig:a}\includegraphics[width=\textwidth, trim = 0 0 0 150,clip ]{images/bmvc_qualitative.pdf}}
\subfigure[\textbf{Failure case}. The initial negative clicks fail to remove the pixels in the body of the doll as the network interprets both the head and the body as a single object. Hence, the network needs more clicks to produce the desired result.]{\label{fig:b}\includegraphics[width=\textwidth]{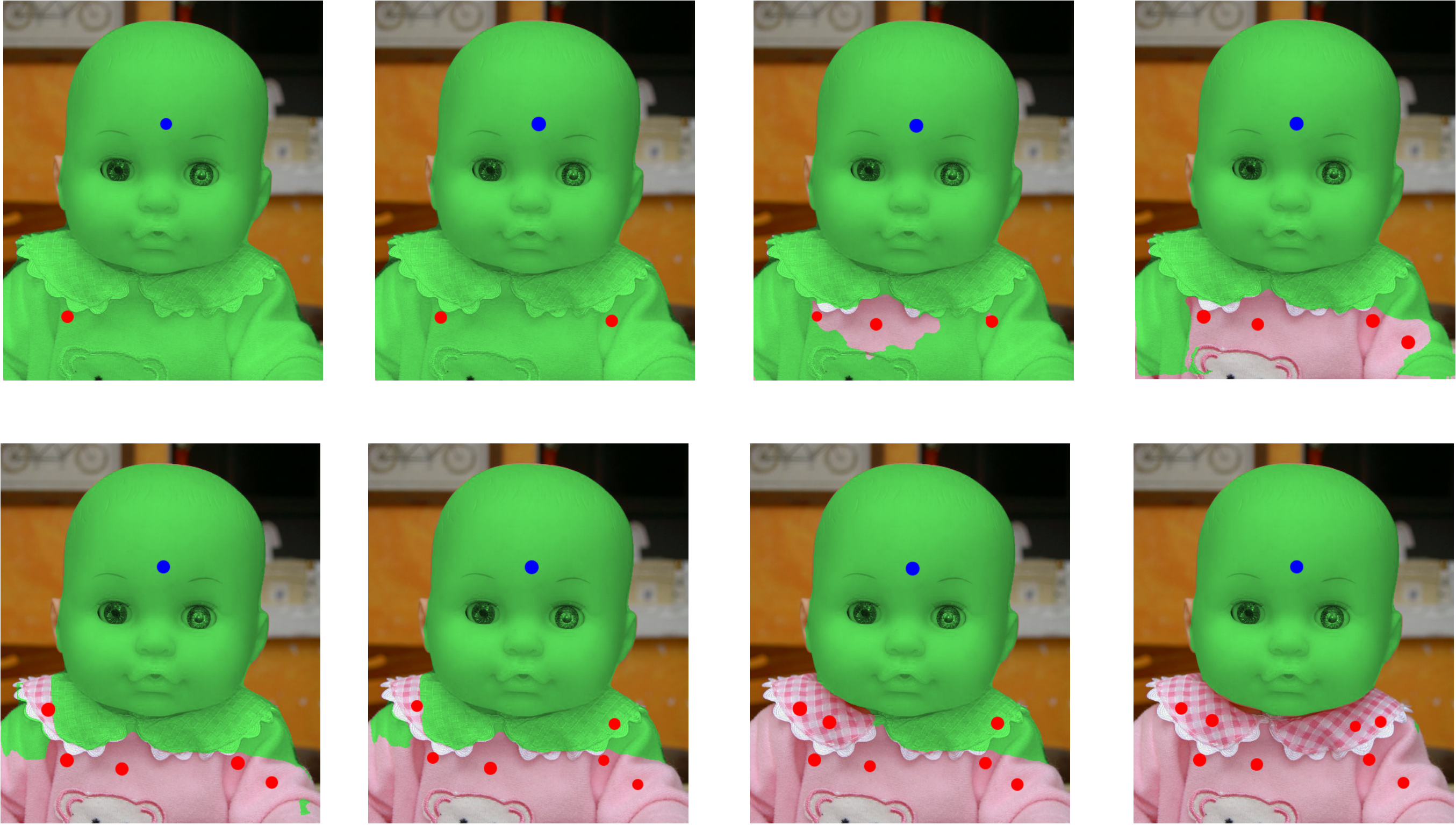}}
\caption{\label{fig:qualitative}Qualitative results of the proposed iteratively trained interactive segmentation method.}
\end{figure}

Figure \ref{fig:qualitative} shows qualitative results of our method.

\end{document}